\documentclass[journal]{IEEEtran}

\ifCLASSOPTIONcompsoc
  \usepackage[nocompress]{cite}
\else
\fi

\ifCLASSINFOpdf
   \usepackage[pdftex]{graphicx}
   \DeclareGraphicsExtensions{.pdf,.jpeg,.png}
\else
\fi

\usepackage[cmex10]{amsmath}
\usepackage{array}

\usepackage{mdwmath}
\usepackage{mdwtab}
\usepackage{eqparbox}

\usepackage{xspace}
\usepackage{times}
\usepackage{graphicx}
\usepackage{amsmath}
\usepackage{amssymb}
\usepackage{algorithmic,algorithm}
\usepackage{yhmath}
\usepackage{url}
\usepackage{bm}
\usepackage{multirow}

\makeatletter
\DeclareRobustCommand\onedot{\futurelet\@let@token\@onedot}
\def\@onedot{\ifx\@let@token.\else.\null\fi\xspace}

\def\eg{\emph{e.g}\onedot}

\def\etc{\emph{etc}\onedot}

\def\etal{\emph{et al}\onedot}
\def\w{{\boldsymbol w}}
\def\x{{\boldsymbol x}}

\makeatother

\begin{document}

\title{Multiple Kernel Learning in the Primal for Multi-modal Alzheimer's Disease Classification}

\author{
    Fayao Liu,
    Luping Zhou,
 	Chunhua Shen,
    Jianping Yin %
 	\thanks{F. Liu and C. Shen are with School of Computer Science, University of Adelaide,
 	 SA 5005, Australia.
    Email: \{fayao.liu, chunhua.shen\}@adelaide.edu.au.
 	Correspondence should be addressed to C. Shen.}
    \thanks{L. Zhou is with
    University of Wollongong, NSW, 2522 Australia.
 	}
    \thanks{J. Yin is with College of Computer,
    National University of Defense Technology, Changsha, Hunan, 410073, China.
    }
    \thanks{Data used in preparation of this article were obtained from
    the Alzheimer's Disease Neuroimaging Initiative (ADNI) database
    (www.loni.ucla.edu/ADNI).}
    }

\maketitle

\begin{abstract}
To achieve effective and efficient detection of Alzheimer's disease (AD), many machine learning methods have been introduced into this realm.
However, the general case of limited training samples, as well as different feature representations typically makes this problem challenging.
In this work, we propose a novel multiple kernel learning framework to combine multi-modal features for AD classification,
which is scalable and easy to implement.
Contrary to the usual way of solving the problem in the dual space, we look at the optimization from a new perspective.
By conducting Fourier transform on the Gaussian kernel, we explicitly compute the mapping function,
which leads to a more straightforward solution of the problem in the primal space.
Furthermore, we impose the mixed $L_{21}$ norm constraint on the kernel weights, known as the group lasso regularization,
to enforce group sparsity among different feature modalities.
This actually acts as a role of feature modality selection, while at the same time exploiting complementary information among different kernels.
Therefore it is able to extract the most discriminative features for classification.
Experiments on the ADNI data set demonstrate the effectiveness of the proposed method.
\end{abstract}


\section{Introduction}

As the most common type of dementia among the elders, Alzheimer's disease (AD) is now affecting millions of people all over the world.
It is characterized by progressive brain disorder that damages brain cells, leading to memory loss, confusion and eventually to death.
The huge price of caring AD patients has made it one of the most costly diseases in the developed countries,
and also caused great physical, as well as psychological burdens on the caregivers.
From this perspective, early diagnosis of AD can be of great significance.
Identified in an early stage, the disease can be made well under control.

Previous diagnosis mainly depends on evaluation of the patient history, clinical observation, or cognitive assessment.
Recent AD related research showed promising prospect in finding reliable biomarkers for automatic early detection \cite{jieping2011},
which is a promising yet challenging task.
Many projects such as ADNI \cite{website:ADNI} have been launched, to collect data of candidate biomarkers to promote the development of AD research.
Several biomarkers have been studied and proved to be sensitive to Mild Cognitive Impairment (MCI) - an early stage of AD,
\eg, brain atrophy detected by imaging \cite{biomarkerMRI2010},
protein changes in blood or spinal fluid \cite{biomarkerCSF2010},
genetic variations (mutations) \cite{genome2010} \etc.
With accurate early diagnosis of MCI, the progression of converting to AD can be possibly slowed down and well controlled.

Recent studies \cite{Casanova2011,Tripoliti2011} %
indicate that image analysis of brain scans is more reliable and sensitive
in detecting the presence of early AD than traditional cognitive evaluation.
In this context, many machine learning methods have been introduced to perform neuroimaging analysis for automatic AD classification.
Early attempts mainly focused on applying off-the-shelf tools in statistical machine learning to differentiate AD,
with the most popular one being support vector machines (SVMs).

Kl{\"o}ppel \etal \cite{Kloeppel2008} trained a linear SVM to classify AD patients and cognitively normal individuals
using magnetic resonance imaging (MRI) scans. More SVM based approaches can be found in \cite{Fan2008,Shen2012}. %
Besides SVMs, other learning methods are also introduced.
Tripoliti \etal \cite{Tripoliti2011} applied Random Forests on functional MRI (fMRI) obtained from 41 subjects to differentiate AD and health control.
In \cite{Casanova2011}, Casanova \etal implemented a penalized logistic regression to
classify sMRI images of cognitive normal subjects and AD patients from ADNI datasets.
Note that they all used single feature modality for classification.

However, as indicated by \cite{biomarkerCSF2010},
different biomarkers may carry complementary information.
Therefore combining multi-modal features, instead of depending on one is a promising direction for improving classification accuracy.
Intuitively, one can combine multiple results from different classifiers with voting technique, or ensemble method.
Dai \etal \cite{Dai2012} proposed a multi-classifier fusion model through weighted voting,
using maximum uncertainty linear discriminant analysis (MLDA) as base classifiers, to distinguish AD patients and healthy control.
They used features from both sMRI and fMRI images.
Polikar \etal \cite{Polikar2010} proposed an ensemble method based on multi-layer perceptron to combine
Electroencephalography (EEG), positron emission tomography (PET) and MRI data.
A linear program boosting (LP Boosting) algorithm was proposed by Hinrichs \cite{Hinrichs2009a} to jointly consider features
from MRI and fluorodeoxyglucose PET (FDG-PET).

Moreover, concatenating several features into one single vector and then training a classifier can also be a practical option.
Walhovd \etal \cite{Walhovd2010a} performed logistic regression analysis by concatenating MRI, PET and cerebrospinal fluid (CSF) features.
However, such concatenation requires proper normalization of features extracted from different sources;
otherwise the prediction score would be easily dominated by a single feature.
One more disadvantage of this method is that it treats multiple features equally,
being incapable of effectively exploring the complementary information provided by different feature modalities.

In addition to the above stated fusion approaches, another method is multiple kernel learning (MKL)
\cite{Lanckriet:2004:SFG:1092907.1093151,jmlr-SonnenburgRSS06},
which works by simultaneously learning the predictor parameters and the kernel combination weights.
The multiple kernels can come from different sources of feature spaces, thus providing a general framework for data fusion.
It has found successful applications in genomic data fusion \cite{Lanckriet:2004:SFG:1092907.1093151}, protein function prediction \cite{Lanckriet2004PSB} \etc.
As for AD data fusion and classification,
Hinrichs \etal \cite{Hinrichs2009} proposed an MKL method, which casts each feature as one or more kernels
and then solves for support vectors and kernel weights using simplex constraints, known as SimpleMKL \cite{simplemkl}.
Cuingnet \etal \cite{Cuingnet2011} evaluated ten methods for predicting AD,
including linear SVM, Gaussian SVM, logistic regression, MKL \etc, also based on SimpleMKL.
More recently, Zhang \etal \cite{Zhang2011} proposed an SVM based model to combine kernels from MRI, PET and CSF features.
Their formulation does not involve kernel coefficients learning. Instead, they use grid search to find kernel weights, which can be very time consuming or even intractable when the number of kernels or features gets large.
It is worth noting that they all solve the MKL problem in the Lagrange dual space.
Therefore the time complexity scales at least $O(n^{2.3})$ \cite{MKLtimecomplexity} with respect to the size $n$ of the training set.

Here, we propose to directly solve the primal MKL problem.
This is achieved by explicitly computing the mapping function through Fourier extension of the kernel function, inspired by the random features proposed by Rahimi \cite{Rahimi07randomfeatures}.
By sampling components from the Fourier space of the Gaussian kernel using Monte Carlo methods,
we can obtain an approximate embedding, and hence reduce the complexity of the kernel learning problem to $O(n)$.
Furthermore, instead of the most commonly used $L_1$, $L_2$ norm, we impose the mixed $L_{21}$ norm constraint on the kernel weights,
known as the group Lasso, to enhance group sparsity among different feature modalities.
In summary, we highlight the main contributions of this work as follows:
\begin{enumerate}
    \item
We use random Fourier features (RFF) to approximate Gaussian kernels, leading to the straightforward primal solution of the MKL problem.
Therefore the learning complexity is reduced to linear scale.
    \item
We enforce an $L_{21}$ norm constraint on the kernel weights, to promote group sparsity among different feature modalities,
while simultaneously exploiting the complementary information among different kernels.
It can be used to select the most discriminative features to improve classification accuracy.
    \item
The proposed RFF$+L_{21}$ norm MKL framework is used to perform feature selection on ROI feature of AD datasets,
therefore identifying brain regions that are most related to AD. The proposed method yields a simple primal solution
and provides a general framework for heterogeneous feature integration.
\end{enumerate}

The rest of the paper is organized as follows.
Section \ref{sec:methods} first briefly reviews some preliminaries of SVMs and MKL,
and then gives our formulation and the detailed algorithm.
Experimental results are reported and discussed in Section \ref{sec:results}, and conclusions are made in Section \ref{sec:conclusion}.

\section{Methods}
\label{sec:methods}
Before getting into the details of the method, we first define some notation.
A column vector is denoted by a bold lower-case letter ($\mathbf{x}$) and a matrix is represented by a bold upper-case letter ($\mathbf{X}$).
$\mathbf{\xi} \succeq \mathbf{0}$ indicates all elements of $\mathbf{\xi}$ being non-negative.

\subsection{MKL Revisit}
Support Vector Machines (SVMs) \cite{Cortes95support-vectornetworks} is a large margin method, based on the theory of structural risk minimization.
In case of binary classification, SVMs finds a linear decision boundary that best separates the two classes.
When it comes to non-linear separable cases, a mapping function $\Phi:{\mathbb R}^d\rightarrow{\mathbb R}^{d'} (d'>d)$ is adopted to embed the
original data into a higher dimensional space, finally yields linear decision boundary $f(\x)=\w^T\Phi(\x)+b$.
Given a labeled training set $\{(\bm{x}_i,y_i)\}_{i=1}^n$, where $\bm{x}_i \in {\mathbb R}^d$ denotes the training sample
and $y_i \in \{-1,+1\}$ the corresponding class label, canonical SVM solves the following problem:
    \begin{eqnarray} \centering
    {
        \begin{split}
    		\mathop{\min}_{\w,b} \;\; & \frac{1}{2} \| \w \| ^{2} + C\sum_{i }\xi_i \\
    		{\rm s.t.}  \;\; & y_i(\langle \w, \Phi(\x_i) \rangle+b) \ge 1-\xi_i,  \forall{i},	 \\
    		&  \bm {\xi} \succeq \bm{0},  \\
        \end{split}
    }
	\end{eqnarray}
where C is a trade-off parameter between training error and margin maximization,
$\bm{\xi}=[\xi_1,\ldots,\xi_n]^T$ the slack variables, and $\langle \cdot,\cdot \rangle$ represents inner product.
While finding the appropriate mapping function $\Phi$ is always difficult, one usually resorts to solving it in the Lagrange dual space
by the kernel trick:
\begin{eqnarray}\centering
    k(\x, \x_i)= \langle \Phi(\x), \Phi(\x_i) \rangle.
\end{eqnarray}
As $\Phi(\cdot)$ only appears in the inner product form, by such a simple substitution,
one can instead solve the following Lagrange dual problem (3) without explicitly knowing the embedding $\Phi$:
        \begin{align}
    		\mathop{\max}_{\bm{\alpha}}
                \;\; & \sum_i \alpha_i -
                    \frac{1}{2} \sum_{i,j}\alpha_i \alpha_j y_i y_j k(\x_i, \x_j)
                    \notag
                    \\
    		{\rm s.t.}  \;\; & 0\leq \alpha_i \leq C,  \forall{i}; \;\;
    		 \sum_i \alpha_i y_i = 0.
                     \label{f:svmdual}
        \end{align}
Here $\alpha_i$ are Lagrange multipliers, and $k$ the kernel, which is typically predefined.
Several frequently involved kernels are linear, polynomial, Gaussian, sigmoid kernel \etc.

To this end, the algorithm performance relies largely on the kernel one chooses.
While finding the appropriate kernel may not be straightforward,
many researchers turned to using multiple kernels instead of a single one and tried to find the optimum combination of them.
The different kernels may correspond to different similarity representations or different feature sources.
A simple option is to consider the convex combination of basic kernels:
    \begin{eqnarray} \centering \label{formula:f_kernelcomb}
    {
        \begin{split}
    		&k(\x_i, \x_j) = \sum_{m} \beta_m k_m(\x_i, \x_j)  \\
        \end{split}
    }
	\end{eqnarray}
with $ \sum_{m} \beta_m = 1, \bm{\beta} \succeq  \bm{0}$, where $\beta_m$ denotes the weight of the $m$th kernel function.

The process of learning the kernel weights while simultaneously minimizing the structural risk is known as the multiple kernel learning (MKL).
As one of the state-of-the-art MKL algorithms, SimpleMKL \cite{simplemkl} efficiently solves a simplex constrained MKL formulation.
The primal MKL problem with $L_1$ norm constraint is formulated as:
    \begin{eqnarray} \centering \label{formula:f_smkl}
    {
        \begin{split}
    		\mathop {\min }_{\w,\bm{\beta}, \bm{\xi} } \;\;
            &  \frac{1}{2}  \sum_{m}  \frac{1}{\beta_{m}} \|\w_{m}\|^{2}_{2} + C\sum_{i}\xi_i  \\
    		{\rm s.t.} \; \; & y_i(\w^T \sum_l  \Phi_l(\x_i)+b) \ge 1-\xi_i,  \forall{i}  \\
    		& \sum_m \beta_m = 1,  \\
    		& \bm{\beta} \succeq \bm{0}, \bm{\xi} \succeq \bm{0},  \\
        \end{split}
    }
	\end{eqnarray}

While the $L_1$ norm is known as a sparsity inducing norm, one can easily replace the simplex constraint $\sum_m \beta_m = 1$ with the ball constraint $\sum_m \beta_m ^2  \le 1$,
which usually yields the non-sparse solution.
Again, the mapping $\Phi$ is conducted implicitly, which draws its corresponding Lagrange dual problem into spotlight:
    \begin{eqnarray} \centering \label{formula:f_smkldual}
    {
        \begin{split}
		\mathop{\min}_{\bm{\beta}} \mathop{\max}_{\bm{\alpha}} \;\; & \sum_{i}\alpha_i - \frac{1}{2} \sum_{i,j}\alpha_i\alpha_j y_i y_j \sum_{m} \beta_m k_m(\x_i,\x_j) \\
    	   {\rm s.t.} \;\; & \sum_i \alpha_i y_i = 0,  \\
    	   &  0 \le \alpha_i \le C,  \forall{i}, \\
    	   & \sum_m \beta_m = 1, \bm{\beta} \succeq \bm{0},
        \end{split}
    }
    \end{eqnarray}
where $\alpha_i,\alpha_j$ are Lagrange multipliers and $k_m(\x_i,\x_j)$ is the $m$th kernel function.

\subsection{Proposed MKL for Combining multi-modal features}
 MKL provides a principled way of incorporating multi-modal features by using multiple kernels. However, due to the unknown mapping $\Phi$, they usually must be solved in the Lagrange dual space, which results in a time complexity of at least $O(n^{2.3})$ \cite{MKLtimecomplexity} with respect to the data size $n$. We thus seek to look at the MKL problem from a new perspective. Instead of solving it in the dual space, we propose to directly approximate the mapping function through Fourier transform of the kernels,
 leading to the primal solution of the problem. This is originally inspired from the random features proposed by Rahimi \cite{Rahimi07randomfeatures}.
 Specifically, we explicitly seek a $\Psi(\cdot)$ satisfying
 \begin{eqnarray} \centering \label{formula:f_kernelapp}
    k(\x_i, \x_j) \approx \langle \Psi(\x_i), \Psi(\x_j)\rangle
 \end{eqnarray}
Therefore we can simply transform the primal data with $\Psi$ and solve the primal MKL problem in the new feature space.
In this section, we will first introduce the random Fourier features, and then give our formulation and the detailed algorithm.

\subsubsection{Random Fourier Features (RFF)}
In order to approximate $\Phi$, we conduct Fourier transform on kernel functions.
Here, we adopt the most commonly used Gaussian kernel, whose Fourier transform \cite{Rahimi07randomfeatures} is illustrated in Table. \ref{tab:11}.
As can be seen from the table, the Fourier transform of a Gaussian function also conforms to a Gaussian distribution.
Moreover, the bandwidth $\sigma$ in time space corresponds to $\frac{1}{\sigma}$ in Fourier frequency space.
Therefore, we can adopt random Fourier basis $\cos(\bm{\omega}' \x)$ and $\sin(\bm{\omega}' \x)$
to represent the random feature mapping $\Psi$, where $\bm{\omega} \in {\mathbb R}^d$,
are random variables drawn from frequency space of Gaussian kernel using Monte Carlo sampling.
\begin{table}
\centering
\caption{Gaussian kernel and its corresponding Fourier transform} \label{tab:11}
\vspace{-.2cm}
\begin{tabular}{ c c c }
    \hline \hline
    kernel name & k(t) & p($\omega$)\\
    \hline
    Gaussian  & $e^{- \frac{t^2}{2\sigma^2}}$  & $\sqrt{2\pi}\sigma e^{-\frac{\omega^2\sigma^2}{2}}$\\
    \hline \hline
   \vspace{-.7cm}
\end{tabular}
\end{table}

The algorithm of computing random feature map $\Psi$ can be described as Algorithm. \ref{algo:RFF}:

    \begin{algorithm} [htb]
    \caption{Compute random Fourier feature} \label{algo:RFF}
    \begin{algorithmic}
          \STATE \hspace{-1.3em}
          \vspace{.7mm}
          \small{
                {\bf Input:} {Matrix of training samples $\mathbf{X}$, Fourier size $D$, Gaussian kernel bandwidth $\mathbf{\sigma}$} \\
                1. Compute gaussian kernel matrix $\mathbf{K}$.\\
                2. Compute the Fourier transform $p$ of the kernel.\\
                3. Draw $D$ samples $\bm{\omega}_1, \bm{\omega}_2, \ldots, \bm{\omega}_D \in {\mathbb R}^d $ from $p$ by Monte Carlo sampling.\\
                4. $\Psi(\mathbf{X}) = \frac{1}{\sqrt{D}}[\cos(\bm{\omega}_1' \mathbf{X}), \ldots, \cos(\bm{\omega}_D' \mathbf{X}), \sin(\bm{\omega}_1' \mathbf{X}), \ldots,$ \\
                \quad \quad \quad \quad \quad \quad \quad $\sin(\bm{\omega}_D' \mathbf{X})]$ \\
                {\bf Output:} $\Psi(\mathbf{X})$
          }
    \end{algorithmic}
    \end{algorithm}

\subsubsection{Proposed MKL Framework}
Given $p$ different feature groups, the samples are represented as $\mathbf{X}=\{(\x_i^{(1)},\ldots, \x_i^{(p)})\}_{i=1}^N$.
For each feature group, we use $q$ kernel functions to produce $q$ embeddings.
After explicitly computing the random fourier features $\Psi$ according to each kernel, we propose to solve the following primal objective function:
\begin{eqnarray} \centering \label{formula:f_mklalgo}
{
    \begin{split}
    		\mathop{\min}_{\w, \bm{\beta},\bm{\xi}} \;\;
    & \frac{1}{2} \sum_{l=1}^p \sum_{m=1}^q  \frac{1}{\beta_{lm}} \|\w_{lm}\| ^{2} + C\sum_{i=1}^N\xi_i \\
    		{\rm s.t.} \;\; & y_i(\sum_{l=1}^p \sum_{m=1}^q \w^T_{lm} \Psi_{lm}  (\x_i^{(l)})+b) \ge 1-\xi_i,  \forall{i}  \\
    		& \sum_{l=1}^p \|\bm{\beta}_l  \|_2 \le 1, \\
    		& \bm{\beta} \succeq \bm{0}, \bm {\xi} \succeq \bm{0}, \\
    \end{split}
}
\end{eqnarray}
where $l$ indexes different feature groups and $m$ indexes multiple kernels used for a single feature group.
This is a convex optimization problem, which can be efficiently solved using off-the-shelf solvers
like CVX \cite{Boyd:2004}, MOSEK \cite{MOSEK}.

It is worth noting that we use the well known group Lasso ($L_{21}$ norm) constraint of the kernel weights instead of the commonly used $L_1$ norm.
As according to Yan et al. \cite{Yan_Mikolajczyk_Kittler_Tahir_2009}, the $L_1$ norm is less effective
when the combined kernels carry complementary information.
While as stated above, different biomarkers of AD may carry complementary knowledge,
which serves as a reason why the $L_1$ norm underperforms other formulations,
as indicated by experiments later.
Instead, the mixed $L_{21}$ norm formulation enforces group sparsity among different feature modalities,
which actually performs as a role of feature modality selection,
while at the same time exploiting complementary information among the different kernels.
Note that this group Lasso constraint has been widely used and proved to be of great success \cite{GLMKL1,GLMKL2}.
To demonstrate the effectiveness of the proposed RFF$+L_{21}$ norm framework, we also implemented the RFF$+L_1$, RFF$+L_2$ norm formulation,
simply by substituting the $\sum_{l} \|\bm{\beta}_l  \|_2 \le 1$ constraint to $\|\bm{\beta}\|_1 \le 1$,
$\|\bm{\beta}\|_2 \le 1$, respectively.
The decision function thus can be written as
\begin{eqnarray} \centering \label{formula:f_decfunc}
    f(\x) = {\rm sign} (\sum_{l=1}^p \sum_{m=1}^k \w^T_{lm} \Psi_{lm}(\x^{(l)})+b )
\end{eqnarray}

The overall framework is described in Algorithm. 2:
    \begin{algorithm} [htb]
    \caption{Proposed MKL Algorithm} \label{algo:RFFMKL }
    \begin{algorithmic}
          \STATE \hspace{-1.3em}
          \vspace{.7mm}
          \small{
                {\bf Input:} {Training samples $\{({\x}_{i}^{(l)},y_i)\}_{i=1}^N$, trade-off parameter $C$, Gaussian kernels $\mathbf{K}_{lm}$, Fourier size $D$}\\
                1. {\bf for} each kernel matrix $\mathbf{K}_{lm}$ {\bf do}  \\
                \hspace{1.5em} Compute $\Psi_{lm}$ by Alg. \ref{algo:RFF} \\
                2. Solve the primal MKL formulation (\ref{formula:f_mklalgo}) \\
                {\bf Output:} {$\mathbf{w}_{lm}, b$}
          }
    \end{algorithmic}

    \end{algorithm}
 \vspace{-.4cm}

\section{Results and discussion}
\label{sec:results}

To evaluate the performance of the proposed MKL framework, we conduct experiments on the AD dataset obtained from ADNI \cite{website:ADNI}.
The Fourier transform parameter $D$ in our method is set to 2000, and a 5 fold cross validation is conducted on the training set to optimize C (trying values 0.01, 0.1, 1, 10, 100).
We use Gaussian kernels with ten different
kernel bandwidths ($\{2^{-3}, 2^{-2},..., 2^{6}\}$ multiplied by $\sqrt{d}$ with $d$ being the dimension of the feature) for each feature representation, which yields 40 kernels in total.

\subsection{Subjects and data preprocessing}
The  AD  dataset  is composed of 120 subjects, randomly drawn from the
Alzheimer Disease Neuroimaging Initiative (ADNI) database. It includes
70 healthy controls (HC) and 50 progressive MCI patients (PMCI) that
developed probable AD after the baseline scanning.

Each subject is represented by a 229 dimensional feature, coming from two heterogeneous data sources: cerebrospinal fluid (CSF) biomarkers and magnetic resonance imaging (MRI). We categorize the MRI feature into three groups, namely, left hemisphere
hippocampus shape (HIPL), right hemisphere hippocampus shape (HIPR) and grey matter volumes within Regions of Interest (ROI), as they captures different aspects of information. We refer them (CSF, HIPL, HIPR, ROI) as four feature representations.
For more details, the CSF biomarkers are provided by ADNI, including
baseline CSF Ab (42), total tau (t-tau) and phosphorylated tau
(p-tau (181)). The hippocampal shapes are extracted from T1-weighted
MRI and represented by spherical harmonics (SPHARM) for each
hemisphere. To mitigate the influence of misalignment, a
rotation-invariant SPHARM representation \cite{siggraph2003} is employed, which also
reduces the dimensionality of the shape
descriptors. The brain regional grey matter volumes are measured
within 100 Regions of Interest (ROI) via an ROI atlas \cite{ROIatlas} on tissue
segmented brain images that have been spatially normalized into a
template space \cite{3datlas} after intensity correction, skull stripping, and
cerebellum removal.

We summarize the features in Table. \ref{tabFeature}.
The CSF and ROI features are normalized to 0 means with unit variations.
\begin{table}[!hbp]
\centering
\caption{Four feature representations of the AD dataset.} \label{tabFeature}
\vspace{-.2cm}
    \begin{tabular}{ l c c c }
        \hline \hline
        Name &Dimension &Data Source &Representation  \\
	   \hline
        CSF &3  &CSF  &Cerebrospinal fluid  \\
        HIPL  &63 &MRI  &Left hippocampus shape \\
        HIPR &63  &MRI  &Right hippocampus shape\\
	    ROI &100  &MRI  &ROI volume \\
        \hline %
    \end{tabular}
    \vspace{-.5cm}
\end{table}

\subsection{AD classification}
To give an overall evaluation of the proposed method, in addition to the prediction accuracy (ACC), we use four indicators, namely, sensitivity (SEN), specificity (SPE), Matthews correlation coefficient (MCC)\cite{Matthews1975} and the area under the ROC curve (AUC).

We run the proposed algorithms 20 times on the AD dataset with randomly partitioned training and testing sets (2/3 for training and 1/3 for testing).
The best accuracy results of SVM by using different kernels on each single feature representation and on the concatenated features (denoted as SVM (All))
are used as baselines. Table. \ref{tab:acc} reports the results of mean$\pm$std, with best scores highlighted in bold.
As can be observed, among all the four types of features, ROI feature appears to be the most discriminative one, with an accuracy of 82.63\%.
Combining features from multiple modalities indeed outperforms the best single feature based classifier. Even a simple concatenation can improve the performance.
As indicated by the MCC values, the proposed RFF$ +L_{21}$ formulation achieves the best overall performance,
being slightly better than the SimpleMKL.
The $L_{21}$ norm turns out to be more effective than the $L_1$, $L_2$ norm.

\begin{table*}
\centering
\caption{Comparison of performance using single and multi feature representation classification methods on the AD dataset over 20 individual runs.}  \label{tab:acc}
\vspace{-.2cm}
\begin{tabular}{ l c c c c c}
  \hline \hline
  Method &ACC(\%)  &SEN(\%) &SPE(\%)  &MCC(\%) &AUC \\
  \hline
  SVM (CSF)  &78.38 $\pm$ 5.58 &80.30 $\pm$ 8.13 &75.05 $\pm$ 9.53 &55.43 $\pm$ 11.56  &0.826 $\pm$ 0.064 \\
  SVM (HIPL) &77.75 $\pm$ 5.90 &84.21 $\pm$ 7.38 &69.61 $\pm$ 10.86 &54.16  $\pm$ 12.19 &0.844 $\pm$ 0.059 \\
  SVM (HIPR) &77.50 $\pm$ 6.49  &81.53 $\pm$ 8.24 &72.97 $\pm$ 13.10  &54.30 $\pm$ 13.42  &0.832 $\pm$ 0.069 \\
  SVM (ROI)  &82.63 $\pm$ 5.10 &\textbf{94.02 $\pm$ 5.42} &66.50 $\pm$ 7.24 &64.58 $\pm$ 9.91  &0.899 $\pm$ 0.040 \\
  SVM (All)  &83.62 $\pm$ 6.10 &93.91 $\pm$ 5.22 &69.75 $\pm$ 9.41 &66.87 $\pm$ 10.93  &0.913 $\pm$ 0.034 \\
  SimpleMKL  &85.88 $\pm$ 4.00 &90.53 $\pm$ 6.73 &79.47 $\pm$ 7.24 &70.87 $\pm$ 8.13 &0.934 $\pm$ 0.039 \\
  RFF$+L_1$ &83.12 $\pm$ 6.12  &86.35 $\pm$ 7.98 &78.29 $\pm$ 13.00 &65.32 $\pm$ 13.10 &0.905 $\pm$0.034  \\
  RFF$+L_2$ &85.12 $\pm$ 4.62  &87.97 $\pm$ 6.92 &\textbf{80.83 $\pm$ 12.12} &69.42 $\pm$ 9.90 &0.921 $\pm$ 0.033\\
  RFF$+L_{21}$  &\textbf{87.12 $\pm$ 3.37}  &  91.79 $\pm$ 5.08 &\textbf{80.73 $\pm$ 7.35} & \textbf{73.30 $\pm$ 7.37} &\textbf{0.952 $\pm$ 0.038}  \\
  \hline %
\end{tabular}
\vspace{-.4cm}
\end{table*}

For further validation of the proposed method, we design an extra experiment to compare our framework with \cite{Zhang2011}. We implemented their method by exactly following the description in their paper. To be more precise, a coarse grid search through cross validation is adopted to find the optimal kernel weights and then an SVM is trained (solve e.q.(3)) by the selected kernel combination weights and linear kernels. The SVM is implemented by LIBSVM toolbox \cite{libsvm} with $C=1$, as did in \cite{Zhang2011}. We use the same experimental settings as in \cite{Zhang2011}. Specifically, the whole dataset is equally partitioned into 10 subsets, and each time one subset is chosen as test set and all the rest are for training. This process is repeated 10 times for different partitions to ensure unbiased evaluation. For the implementation of \cite{Zhang2011}, a 10-fold cross validation is performed on the training data in each round to determine the optimal kernel weights $\bm{\beta}$ through a grid search ranging from 0 to 1 at a step size of 0.1. For our method and SimpleMKL, we also fix $C=1$ and use the same kernel settings as above. Table. \ref{tabComp} shows the average performance.

\begin{table}[!hbp]
\vspace{-.5cm}
\centering
\caption{Average performance of different methods on the AD dataset.} \label{tabComp}
\vspace{-.2cm}
\begin{tabular}{ l c c c c}
    \hline \hline
    Method &ACC &SEN &SPE &MCC \\
    \hline
    \cite{Zhang2011} &86.39\% &85.74\% &86.93\% &72.02\% \\
    SimpleMKL &87.06\% &87.89\% &86.68\% &74.57\% \\
    RFF$+L_1$ &81.94\% &83.83\% &78.97\% &63.31\%\\
    RFF$+L_2$ &85.00\% &85.49\% &84.28\% &69.41\% \\
    RFF$+L_{21}$  &\textbf{90.56\%} &\textbf{93.26}\% &\textbf{87.49}\% &\textbf{81.98}\% \\
    \hline %
\end{tabular}
\end{table}

According to Table. \ref{tabComp}, our method outperforms \cite{Zhang2011} and SimpleMKL in terms of all the four criteria. The reasons can be summarized as: 1) Our method uses more powerful Gaussian kernels while \cite{Zhang2011} uses linear kernels;  2) Our formulation can easily incorporate more kernels while \cite{Zhang2011} only uses one kernel for each feature representation; 3) By combining RFF with the $L_{21}$ norm, our method exploits the group sparsity as well as the complementary information among different kernels. As for 2), if more kernels are to be added into \cite{Zhang2011}, a much finer grid search would be required to ensure accuracy, which leads to more time expense or even intractable situation. It is also worth noting that in \cite{Zhang2011}, they have used CSF, MRI as well as PET features for reporting their results.
One more conclusion can be made that the $L_2$ norm always outperforms the $L_1$ norm, which may be explained by the fact that the combined kernels carry complementary information.

\begin{figure*}
    \centering
    \vspace{-.8cm}
    \begin{tabular}{ccc}
    \includegraphics[width=0.24\linewidth, height=0.19\linewidth] {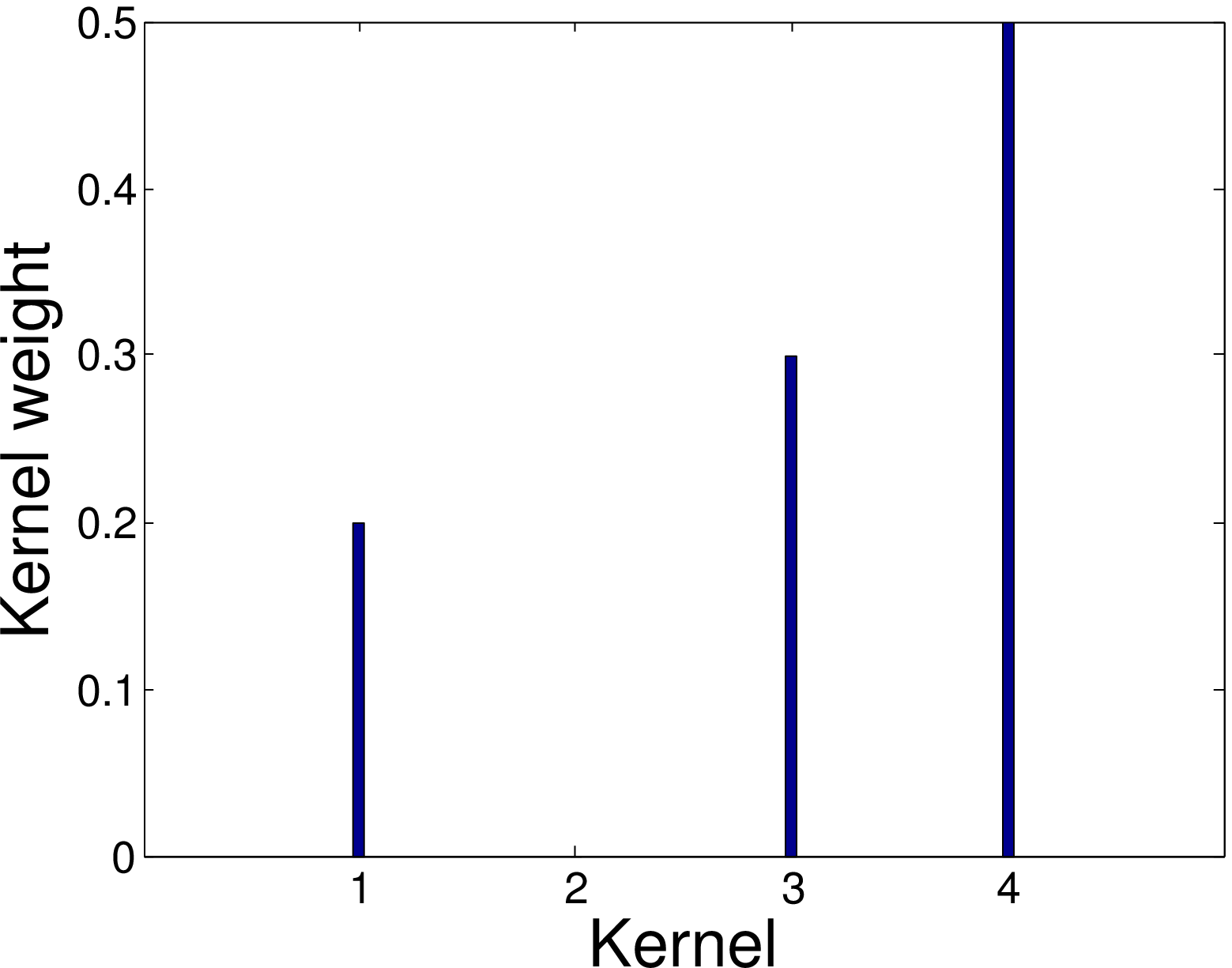} &
    \includegraphics[width=0.24\linewidth, height=0.19\linewidth]{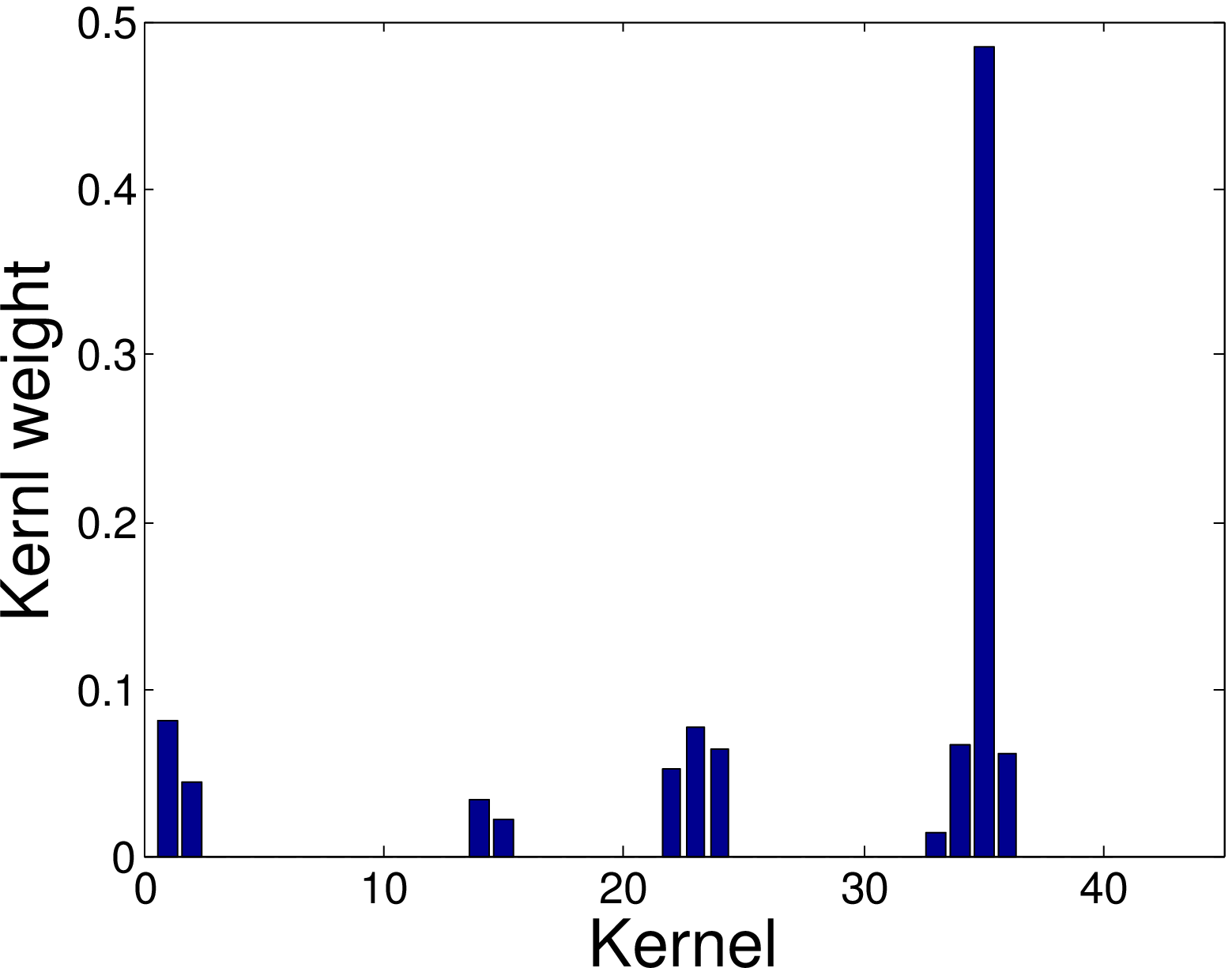} &
    \includegraphics[width=0.24\linewidth, height=0.19\linewidth]{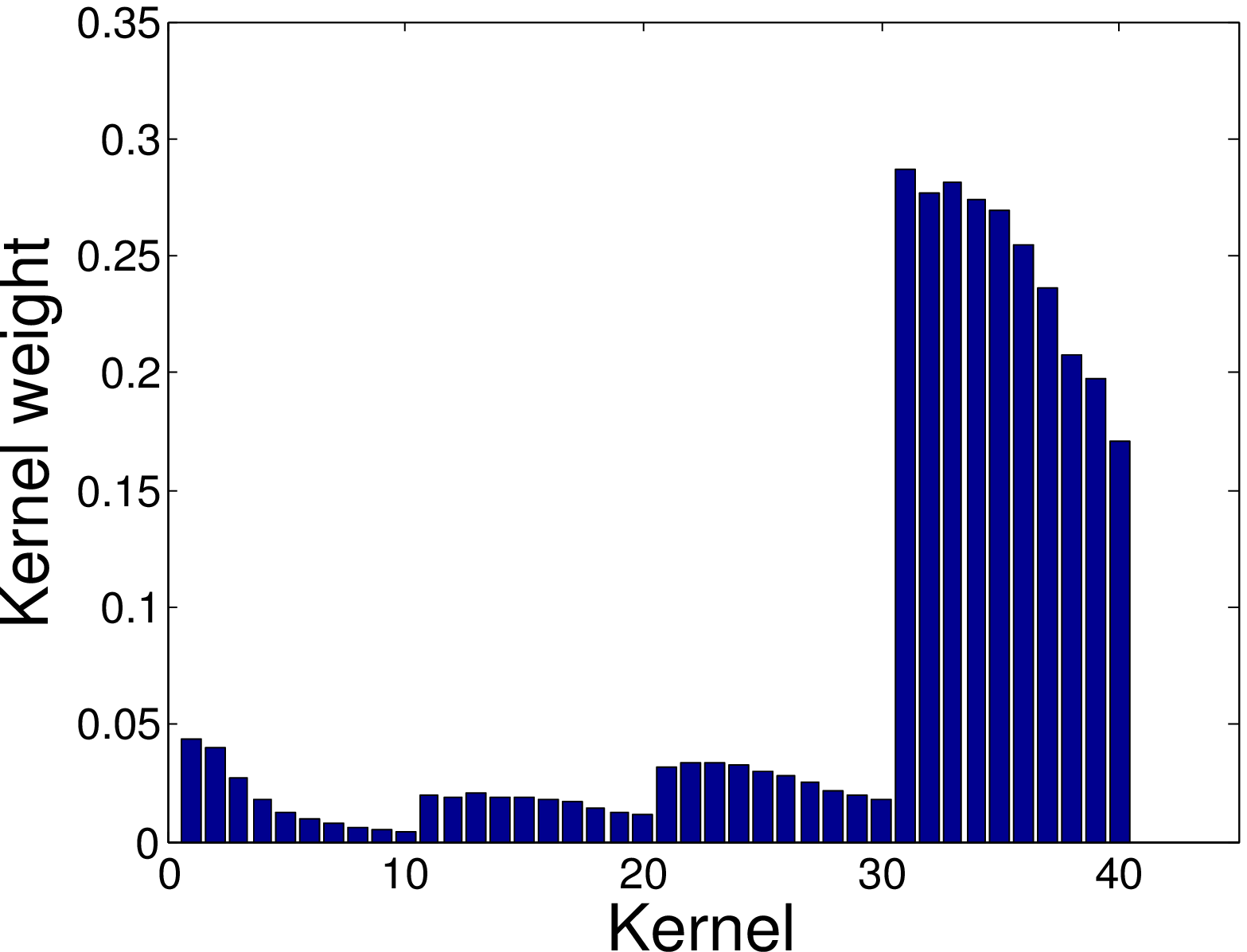} \\
    (a) & (b) & (c)
    \end{tabular}
    \vspace{-.3cm}
    \caption { Base kernel weights comparison of different MKL algorithms on the AD dataset. (a)\cite{Zhang2011}; (b)SimpleMKL; (c)Proposed RFF $+L_{21}$ norm formulation. In (a), according to \cite{Zhang2011}, only one linear kernel is used for each feature representation. In (b) and (c), from left to right, every ten kernels correspond to CSF, HIPL, HIPR, ROI respectively.}
\label{fig:kernelweights}
\vspace{-.3cm}
\end{figure*}

To better illustrate how the multiple kernel methods work, we choose one best performed run for each method and give the kernel weights comparison in Fig. \ref{fig:kernelweights}.
As can be seen, in all the methods, kernels corresponding to the ROI feature are assigned the highest weights.
In other words, they select ROI as the most discriminative feature representation, which is in accordance with the conclusion from single feature based SVM classifier shown in Table. \ref{tab:acc}.

\subsection{Identify brain regions closely related to AD}
In order to identify which areas of the brain region are closely related to AD,
we conduct a further experiment to select the most discriminative ROI features.
As mentioned above, by imposing $L_{21}$ norm constraint on the kernel weights,
group sparsity are enforced, which actually acts as a role of feature selection.
Therefore we can treat each dimension of the ROI (each represents a certain brain region)
as an individual feature to perform the RFF$+L_{21}$ algorithm, leading to sparsity
among different brain regions.
More specifically, we set $p=100$ (group size equals 1) and use
$\mathbf{X}_{ROI}=\{\x_i^{(1)},\x_i^{(2)},\ldots,\x_i^{(p)}\}_{i=1}^N$ as input
to Algorithm. 2, and then rank the regions according to the corresponding kernel weights.

\begin{table} [!hbp]
\vspace{-.5cm}
\centering
\caption{The selected top 20 ROI regions with their corresponding average kernel weights and classification accuracy.} \label{tab:ROIregions}
\vspace{-.2cm}
\begin{tabular}{ l c c}
    \hline \hline
    ROI region &  Kernel weight &ACC (\%) \\
    \hline
    \textbf{hippocampal formation right} &0.1364  &\textbf{75.62}\\
    \textbf{hippocampal formation left} &0.1188  &\textbf{80.57} \\
    \textbf{occipital pole left}  &0.1077 &\textbf{82.75}  \\
    uncus left  &0.1077 &80.68 \\
    \textbf{lateral ventricle right} &0.1029 &\textbf{81.63}   \\
    fourth ventricle right  &0.0803 &80.25  \\
    \textbf{perirhinal cortex left} &0.0782 &\textbf{81.35} \\
    \textbf{amygdala left}  &0.0761 &\textbf{82.38}\\
    \textbf{lateral ventricle left} &0.0517 &\textbf{83.75} \\
    subthalamic nucleus right &0.0493 &83.37  \\
    putamen right  &0.0491 &82.88 \\
    \textbf{inferior frontal gyrus left}&0.0457  &\textbf{84.63}  \\
    middle occipital gyrus right &0.0404  &84.12  \\
    corpus callosum  &0.0391  &84.63    \\
    \textbf{precuneus right} &0.0379  &\textbf{85.75} \\
    \textbf{medial occipitotemporal gyrus right} &0.0373  &\textbf{88.00}\\
    nucleus accumbens left  &0.0372  &87.13 \\
    perirhinal cortex right &0.0362  &87.62 \\
    supramarginal gyrus left &0.0355  &87.87 \\
    medial occipitotemporal gyrus left &0.0327  &87.25 \\
    \hline %
\end{tabular}
\end{table}

For each dimension of the ROI feature, we use three Gaussian kernels $\sigma= \{0.5,1,2\}$ with $C=1$.
We randomly split the dataset into 2/3 for training and 1/3 for testing and report the average performance over 10 different trials.
The selected top 20 regions and their average kernel weights are summarized in Table. \ref{tab:ROIregions}.
Note that the average kernel weights are summed over all different bandwidth kernels.

\begin{figure}
\vspace{-.3cm}
    \centering
    \includegraphics[width=0.35\textwidth]{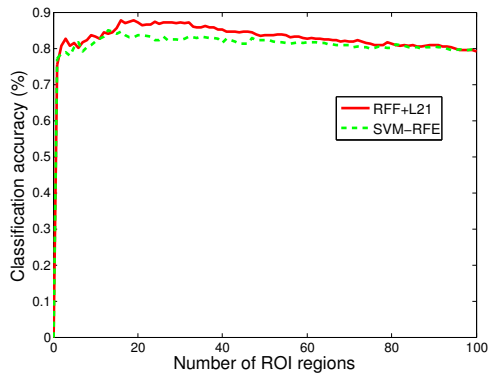}
\vspace{-.3cm}
    \caption {Classification accuracy with respect to different number of selected ROI regions.} \label{fig:acc_roi_sel}
 \vspace{-.6cm}
\end{figure}

To quantitatively evaluate the effect of the feature selection, we test the classification accuracy with respect to different numbers of the selected ROI regions. For a comparison, we also implement an SVM Recursive Feature Elimination method described in \cite{SVM-RFE}, referred as SVM-RFE, which is a popular feature selection method. Then according to the feature rankings, we use an increasing number of ROI features to train a Gaussian SVM with bandwidth $\sigma=\sqrt{d}$ (d is the number of ROI features) and $C=1$. The evaluation is averaged over 20 different runs using 2/3 for training and 1/3 for testing. Fig. \ref{fig:acc_roi_sel} shows the results. As can be seen, using features selected by our method is similar but statistically better than SVM-RFE. Moreover, the classification accuracy of the proposed RFF$+L_{21}$ reaches its peak at the number of 16, and better than using all the ROI regions. We further calculate the pairwise correlations of the top 16 features selected by each method and get the average correlation coefficients of 0.3212 and 0.3661 for RFF$+L_{21}$ and SVM-RFE respectively. This explains the performance in Fig. \ref{fig:acc_roi_sel}, as the features selected by SVM-RFE are more correlated than those selected by RFF$+L_{21}$.
Inspired from this, we use the top 16 ranked ROI regions to reproduce the first experiment and get an accuracy of $90.75\% \pm 3.25$, even better than the one ($87.12\% \pm 3.37$) we reported in Table. \ref{tab:acc}. This further demonstrates the efficacy of the feature selection using the proposed method.

\begin{figure*}
    \centering
    \begin{tabular}{cccc}
    \includegraphics[width=0.23\textwidth]{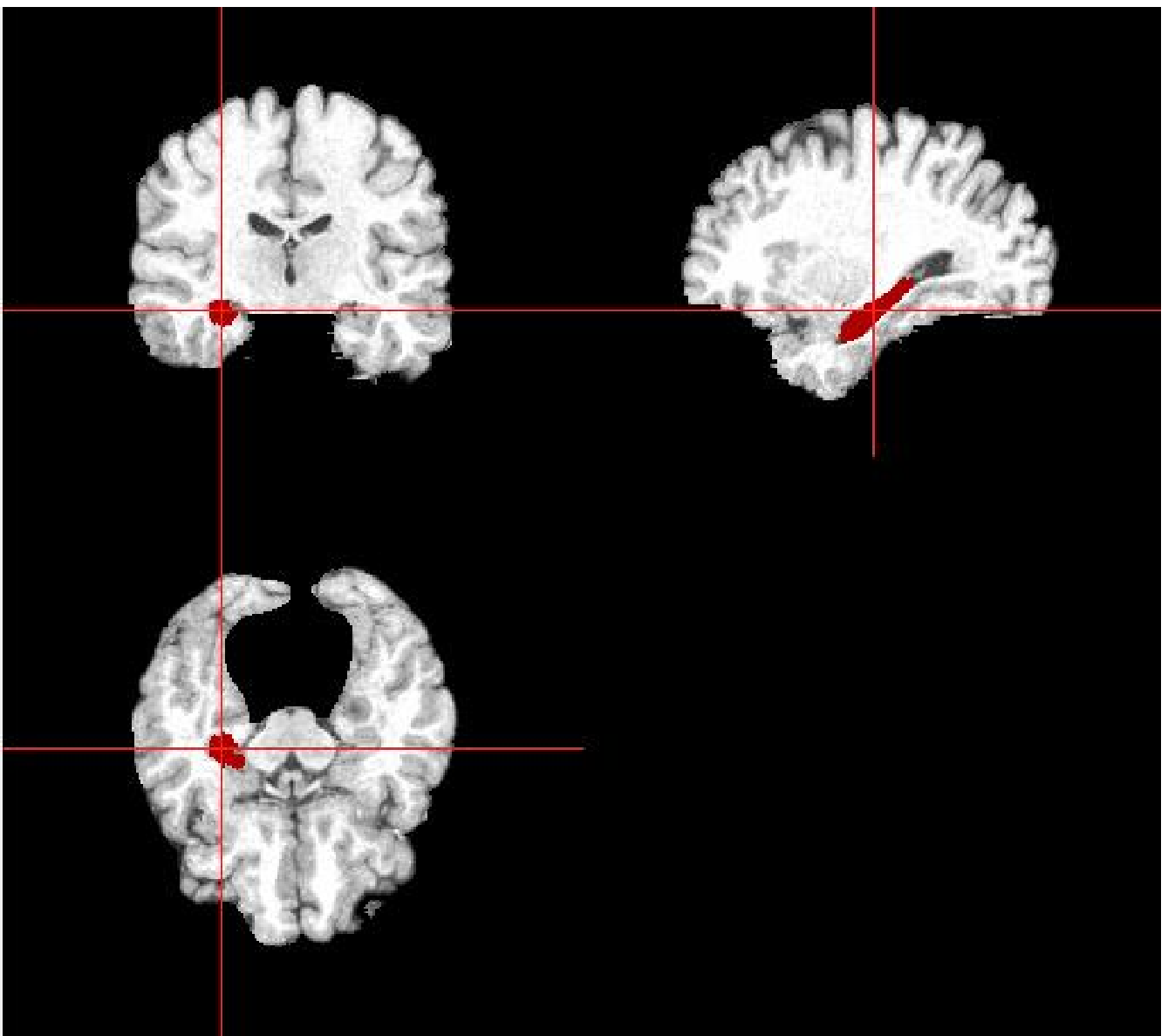}&
    \includegraphics[width=0.23\textwidth]{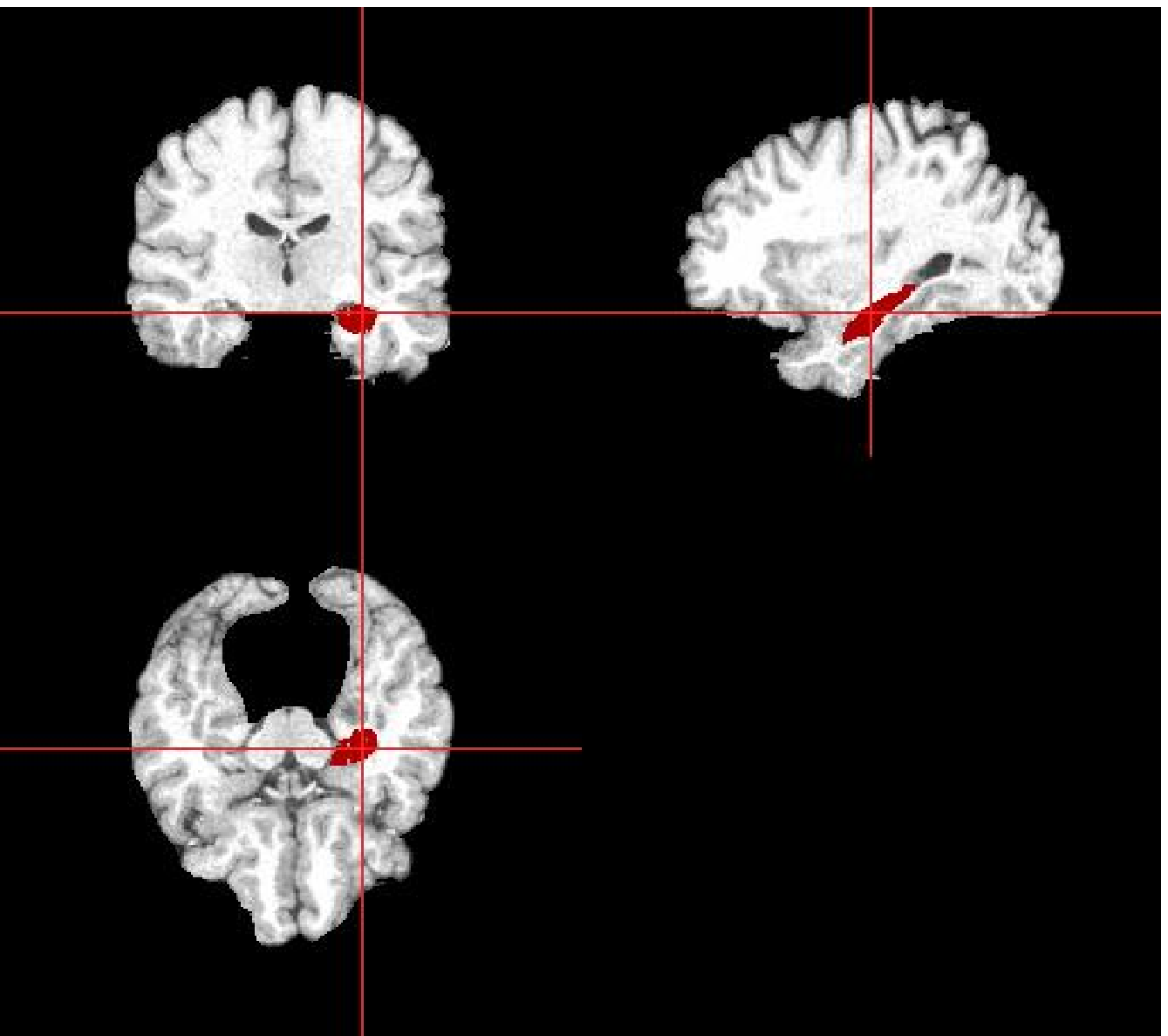}&
    \includegraphics[width=0.23\textwidth]{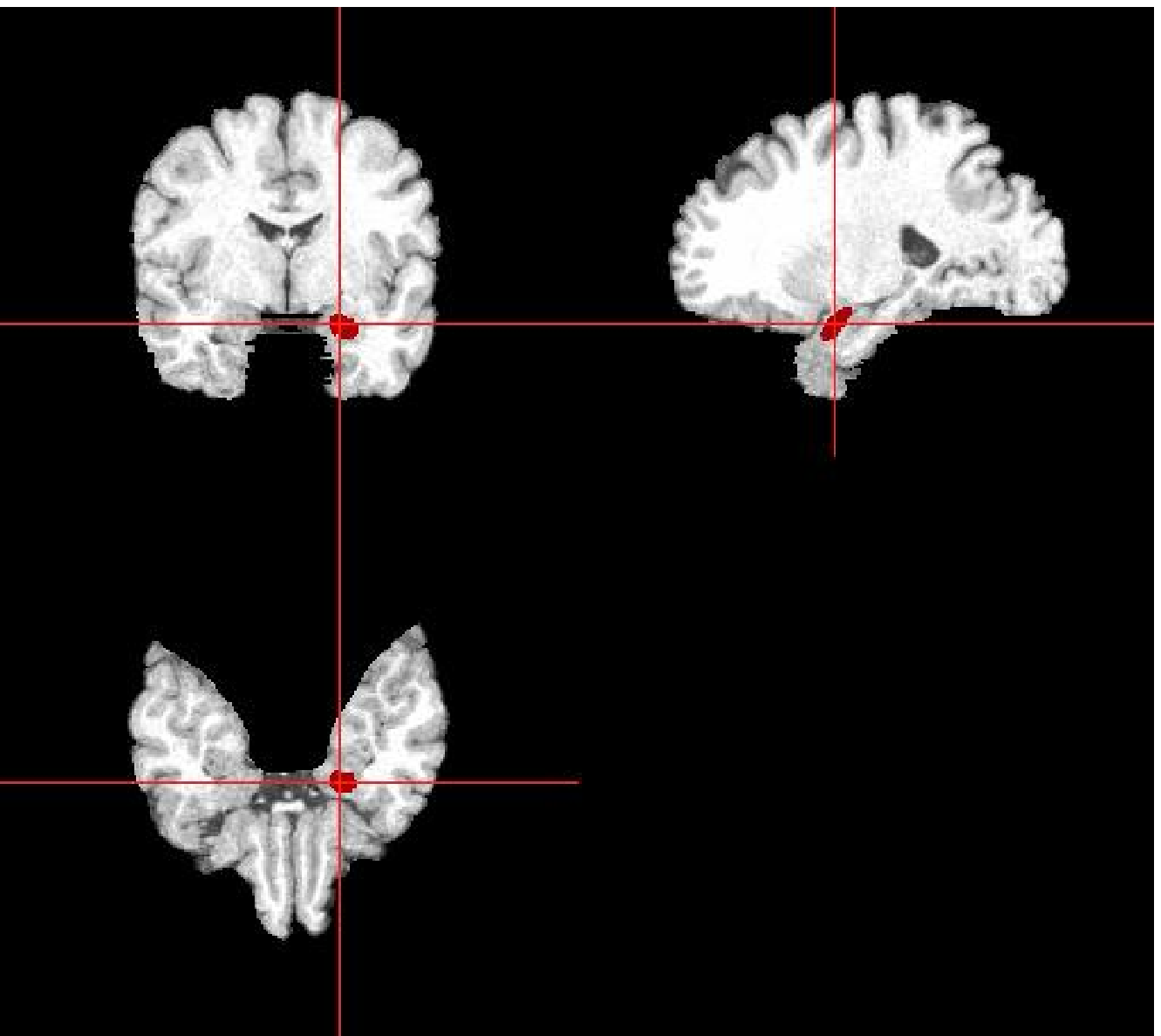}&
    \includegraphics[width=0.23\textwidth]{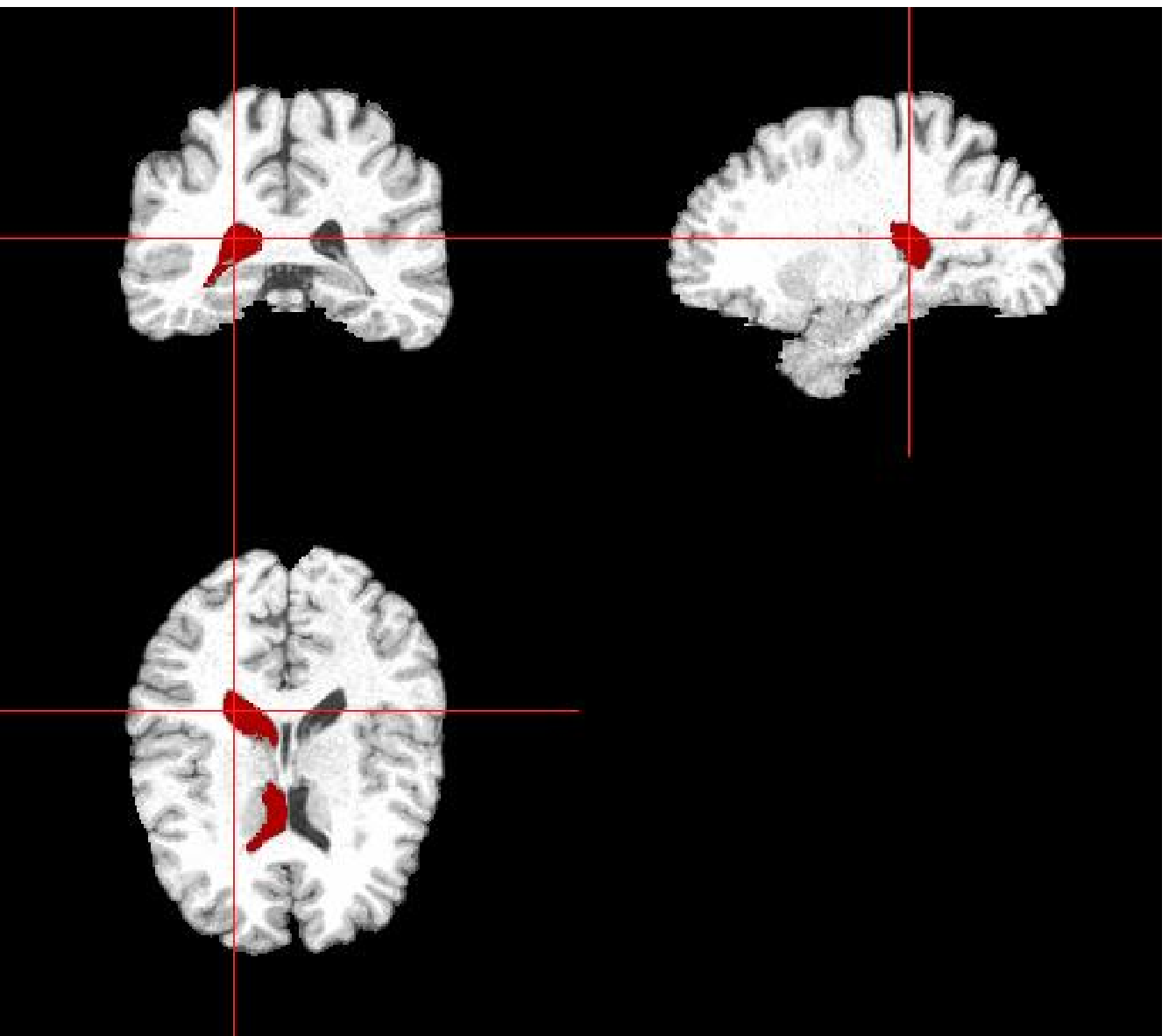}\\
     (a)  & (b) &(c)  & (d)
    \end{tabular}
    \vspace{-.4cm}
    \caption {Four representative brain regions selected by the proposed RFF$+L_{21}$ method.
    (a) hippocampal formation right; (b) hippocampal formation left;
    (c) amygdala left; (d) lateral ventricle right.} \label{fig:ROIregions}
    \vspace{-.5cm}
\end{figure*}

From Fig. \ref{fig:acc_roi_sel}, we can further identify the most discriminative features among the top 20. We list the classification accuracy of the top 20 regions in Table. \ref{tab:ROIregions}. By selecting the one which significantly increases the accuracy according to the curve in Fig. \ref{fig:acc_roi_sel}, we highlight the potential regions closely related to AD in bold. Among them, `hippocampal formation right', `hippocampal formation left', `amygdala left', `precuneus right', `lateral ventricle right', `medial occipitotemporal gyrus' are commonly known to be related to
AD by many studies in the literature \cite{MisraFD09,karas2007pre,amygdala}.
As examples, hippocampus, a brain area closely related to the memory, is especially vulnerable and always affected in the occurrence of AD \cite{MisraFD09};
in \cite{amygdala}, agymdala atrophy was claimed comparable to hippocampal atrophy in AD patients;  precuneus atrophy was observed in early-onset of AD in \cite{karas2007pre}. Fig. \ref{fig:ROIregions} visualizes four examples of the selected regions (in red) against the atlas MRI with cerebellum removed.

\section{Conclusions}
\label{sec:conclusion}

We have proposed a general but simple multiple kernel learning framework for the AD classification problem by combining multi-modal features.
Instead of solving the problem in the dual space as one commonly does, we propose to explicitly compute the mapping function
through Fourier transform and random sampling, leading to the primal solution of the problem. The proposed method is easy to implement
and scales as the linear time of the sample size.
Also, we impose group Lasso constraint on the kernel weights, to enhance group sparsity among different feature representations,
which selects the most discriminative feature groups, while at the same time exploiting the complementary information among different kernels
within a group.
Experimental results on the AD dataset demonstrate that the proposed RFF+$L_{21}$ norm algorithm outperforms other feature fusion methods.
We further utilize the feature selection of the proposed framework to extract the most discriminative ROI features,
hence identifying brain regions that most related to AD. Conclusions are in accordance with studies in the literature.

%

    %
%\vspace{-.3cm}
{
\small
\bibliographystyle{ieee}
\bibliography{ADref}

\begin{thebibliography}{10}\itemsep=-1pt

\bibitem{website:ADNI}
ADNI.
\newblock Alzheimer disease neuroimaging initiative.
\newblock \url{http://adni.loni.ucla.edu/}, 2011.

\bibitem{GLMKL1}
F.~R. Bach.
\newblock Consistency of the group lasso and multiple kernel learning.
\newblock {\em J. Mach. Learn. Res.}, 2008.

\bibitem{Boyd:2004}
S.~Boyd and L.~Vandenberghe.
\newblock {\em Convex Optimization}.
\newblock 2004.

\bibitem{Casanova2011}
R.~Casanova, C.~T. Whitlow, B.~Wagner, J.~Williamson, S.~A. Shumaker, J.~A.
  Maldjian, and M.~A. Espeland.
\newblock High dimensional classification of structural {MRI} alzheimer's
  disease data based on large scale regularization.
\newblock {\em Front Neuroinform}, 2011.

\bibitem{libsvm}
C.-C. Chang and C.-J. Lin.
\newblock {LIBSVM: A library for support vector machines}.
\newblock {\em ACM Trans. on Intell. Syst. and Technol.}, 2011.

\bibitem{Cortes95support-vectornetworks}
C.~Cortes and V.~Vapnik.
\newblock Support-vector networks.
\newblock {\em Mach. Learn.}, 1995.

\bibitem{Cuingnet2011}
R.~Cuingnet, E.~Gerardin, J.~Tessieras, G.~Auzias, S.~Lehericy, M.~O. Habert,
  M.~Chupin, H.~Benali, and O.~Colliot.
\newblock Automatic classification of patients with alzheimer's disease from
  structural {MRI}: A comparison of ten methods using the {ADNI} database.
\newblock {\em NeuroImage}, 2011.

\bibitem{Dai2012}
Z.~Dai, C.~Yan, Z.~Wang, J.~Wang, M.~Xia, K.~Li, and Y.~He.
\newblock Discriminative analysis of early alzheimer's disease using
  multi-modal imaging and multi-level characterization with multi-classifier.
\newblock {\em Neuroimage}, 2012.

\bibitem{MKLtimecomplexity}
L.~Duan, I.~W. Tsang, and D.~Xu.
\newblock Domain transfer multiple kernel learning.
\newblock {\em {IEEE} Trans. Pattern Anal. Mach. Intell.}, 2012.

\bibitem{Fan2008}
Y.~Fan, S.~M. Resnick, X.~Wu, and C.~Davatzikos.
\newblock Structural and functional biomarkers of prodromal alzheimer's
  disease: a high-dimensional pattern classification study.
\newblock {\em Neuroimage}, 2008.

\bibitem{biomarkerCSF2010}
A.~M. Fjell, K.~B. Walhovd, C.~Fennema-Notestine, L.~K. McEvoy, D.~J. Hagler,
  D.~Holland, J.~B. Brewer, A.~M. Dale, and for~the Alzheimer's Disease
  Neuroimaging~Initiative.
\newblock {CSF} biomarkers in prediction of cerebral and clinical change in
  mild cognitive impairment and alzheimer's disease.
\newblock {\em J. of Neuroscience}, 2010.

\bibitem{biomarkerMRI2010}
G.~B. Frisoni, N.~C. Fox, C.~R. Jack, P.~Scheltens, and P.~M. Thompson.
\newblock The clinical use of structural {MRI} in alzheimer disease.
\newblock {\em Nat. Rev. Neurology}, 2010.

\bibitem{SVM-RFE}
I.~Guyon, J.~Weston, S.~Barnhill, and V.~Vapnik.
\newblock Gene selection for cancer classification using support vector
  machines.
\newblock {\em Mach. Learn.}, 2002.

\bibitem{Hinrichs2009a}
C.~Hinrichs, V.~Singh, L.~Mukherjee, G.~Xu, M.~K. Chung, S.~C. Johnson, and
  A.~D.~N. Initiative.
\newblock Spatially augmented {LP}boosting for {AD} classification with
  evaluations on the {ADNI} dataset.
\newblock {\em Neuroimage}, 2009.

\bibitem{Hinrichs2009}
C.~Hinrichs, V.~Singh, G.~Xu, and S.~Johnson.
\newblock {MKL} for robust multi-modality {AD} classification.
\newblock {\em Medical Image Computing and Computer-Assisted Intervention},
  2009.

\bibitem{3datlas}
N.~Kabani, D.~MacDonald, C.~Holmes, and A.~Evans.
\newblock A 3d atlas of the human brain.
\newblock {\em Neuroimage}, 1998.

\bibitem{karas2007pre}
G.~Karas, P.~Scheltens, S.~Rombouts, R.~van Schijndel, M.~Klein, B.~Jones,
  W.~van~der Flier, H.~Vrenken, and F.~Barkhof.
\newblock Precuneus atrophy in early-onset alzheimer's disease: a morphometric
  structural mri study.
\newblock {\em Neuroradiology}, 2007.

\bibitem{siggraph2003}
M.~Kazhdan, T.~Funkhouser, and S.~Rusinkiewicz.
\newblock Rotation invariant spherical harmonic representation of 3d shape
  descriptors.
\newblock In {\em {ACM} {SIGGRAPH} symp. on Geometry Process.}, 2003.

\bibitem{Kloeppel2008}
S.~Kl{\"o}ppel, C.~M. Stonnington, C.~Chu, B.~Draganski, R.~I. Scahill, J.~D.
  Rohrer, N.~C. Fox, C.~R. Jack, Jr, J.~Ashburner, and R.~S.~J. Frackowiak.
\newblock Automatic classification of {MR} scans in alzheimer's disease.
\newblock {\em Brain}, 2008.

\bibitem{Lanckriet:2004:SFG:1092907.1093151}
G.~R. Lanckriet, T.~De~Bie, N.~Cristianini, M.~I. Jordan, and W.~S. Noble.
\newblock A statistical framework for genomic data fusion.
\newblock {\em Bioinformatics}, 2004.

\bibitem{Lanckriet2004PSB}
G.~R. Lanckriet, M.~Deng, N.~Cristianini, M.~I. Jordan, and W.~S. Noble.
\newblock {Kernel-based data fusion and its application to protein function
  prediction in yeast.}
\newblock {\em Pacific Symp. on Biocomputing.}, 2004.

\bibitem{Matthews1975}
B.~W. Matthews.
\newblock {Comparison of the predicted and observed secondary structure of T4
  phage lysozyme}.
\newblock {\em Biochim. Biophys. Acta}, 1975.

\bibitem{MisraFD09}
C.~Misra, Y.~Fan, and C.~Davatzikos.
\newblock {Baseline and longitudinal patterns of brain atrophy in MCI patients,
  and their use in prediction of short-term conversion to AD: Results from
  ADNI}.
\newblock {\em NeuroImage}, 2009.

\bibitem{MOSEK}
Mosek.
\newblock The {MOSEK} interior point optimizer.
\newblock \url{ http:// www. mosek. com}.

\bibitem{genome2010}
P.~NL.
\newblock Reaching the limits of genome-wide significance in alzheimer disease:
  Back to the environment.
\newblock {\em J. of the American Med. Asso.}, 2010.

\bibitem{Polikar2010}
R.~Polikar, C.~Tilley, B.~Hillis, and C.~M. Clark.
\newblock Multimodal eeg, mri and pet data fusion for alzheimer's disease
  diagnosis.
\newblock {\em {IEEE} Eng. in Med. and Bio. Conf.}, 2010.

\bibitem{amygdala}
S.~Poulin, R.~Dautoff, J.~Morris, L.~Barrett, and B.~Dickerson.
\newblock Amygdala atrophy is prominent in early alzheimer's disease and
  relates to symptom severity.
\newblock {\em Psychiatry Res.}, 2011.

\bibitem{Rahimi07randomfeatures}
A.~Rahimi and B.~Recht.
\newblock Random features for large-scale kernel machines.
\newblock In {\em Proc. Adv. Neural Inf. Process. Syst.}, 2007.

\bibitem{simplemkl}
A.~Rakotomamonjy, F.~R. Bach, S.~Canu, and Y.~Grandvalet.
\newblock {SimpleMKL}.
\newblock {\em J. Mach. Learn. Res.}, {2008}.

\bibitem{ROIatlas}
D.~Shen.
\newblock Very {High-Resolution} morphometry using {Mass-Preserving}
  deformations and {HAMMER} elastic registration.
\newblock {\em NeuroImage}, 2003.

\bibitem{Shen2012}
K.-K. Shen, J.~Fripp, F.~Meriaudeau, G.~Chetelat, O.~Salvado, and P.~Bourgeat.
\newblock Detecting global and local hippocampal shape changes in alzheimer's
  disease using statistical shape models.
\newblock {\em NeuroImage}, 2012.

\bibitem{jmlr-SonnenburgRSS06}
S.~Sonnenburg, G.~R{\"a}tsch, C.~Sch{\"a}fer, and B.~Sch{\"o}lkopf.
\newblock Large scale multiple kernel learning.
\newblock {\em J. Mach. Learn. Res.}, 2006.

\bibitem{Tripoliti2011}
E.~E. Tripoliti, D.~I. Fotiadis, and M.~Argyropoulou.
\newblock A supervised method to assist the diagnosis and monitor progression
  of alzheimer's disease using data from an fmri experiment.
\newblock {\em Artif. Intell. in Medicine}, 2011.

\bibitem{Walhovd2010a}
K.~B. Walhovd, A.~M. Fjell, J.~Brewer, L.~K. McEvoy, C.~Fennema-Notestine,
  D.~Hagler, Jr, R.~G. Jennings, D.~Karow, A.~M. Dale, and A.~D.~N. Initiative.
\newblock Combining {MR} imaging, positron-emission tomography, and {CSF}
  biomarkers in the diagnosis and prognosis of {A}lzheimer disease.
\newblock {\em American J. of Neuroradiology}, 2010.

\bibitem{GLMKL2}
Z.~Xu, R.~Jin, H.~Yang, I.~King, and M.~R. Lyu.
\newblock Simple and efficient multiple kernel learning by group lasso.
\newblock In {\em Proc. Int. Conf. Mach. Learn.}, 2010.

\bibitem{Yan_Mikolajczyk_Kittler_Tahir_2009}
F.~Yan, K.~Mikolajczyk, J.~Kittler, and M.~Tahir.
\newblock A comparison of l1 norm and l2 norm multiple kernel {SVM}s in image
  and video classification.
\newblock {\em Int. Workshop on Content Based Multimedia Indexing}, 2009.

\bibitem{jieping2011}
J.~Ye, T.~Wu, J.~Li, and K.~Chen.
\newblock Machine learning approaches for the neuroimaging study of alzheimer's
  disease.
\newblock {\em {IEEE} Computer}, 2011.

\bibitem{Zhang2011}
D.~Zhang, Y.~Wang, L.~Zhou, H.~Yuan, D.~Shen, and A.~D.~N. Initiative.
\newblock Multimodal classification of alzheimer's disease and mild cognitive
  impairment.
\newblock {\em Neuroimage}, 2011.

\end{thebibliography}
}

\end{document}